\documentclass{ecai} 

\usepackage{hyperref}
\usepackage{latexsym}
\usepackage{amssymb}
\usepackage{amsmath}
\usepackage{amsthm}
\usepackage{booktabs}
\usepackage{enumitem}
\usepackage{graphicx}
\usepackage{color}
\usepackage{xspace}

\newcommand{\sys}{PLDA\xspace}

\newcommand{\BibTeX}{B\kern-.05em{\sc i\kern-.025em b}\kern-.08em\TeX}

\begin{document}


\begin{frontmatter}


\paperid{5442} 



\title{Learning After Model Deployment}


\author[A]{\fnms{Derda}~\snm{Kaymak} 
}
\author[B]{\fnms{Gyuhak}~\snm{Kim}
}
\author[C]{\fnms{Tomoya}~\snm{Kaichi}
}
\author[C]{\fnms{Tatsuya}~\snm{Konishi}
}
\author[A]{\fnms{Bing}~\snm{Liu}
\thanks{Corresponding Author. Email: liub@uic.edu.}
} 

\address[A]{University of Illinois Chicago, USA}
\address[B]{Accenture, USA}
\address[C]{KDDI Research, Japan}


\begin{abstract}
In classic supervised learning, once a model is deployed in an application, it is fixed. No updates will be made to it during the application. This is inappropriate for many dynamic and open environments, where unexpected samples from unseen classes may appear. In such an environment, the model should be able to detect these novel samples from unseen classes and learn them after they are labeled. We call this paradigm \textit{\textbf{Autonomous Learning after Model Deployment}} (ALMD). The learning here is \textit{\textbf{continuous}} and involves \textit{\textbf{no human engineers}}. 
Labeling in this scenario is performed by human co-workers or other knowledgeable agents, which is similar to what humans do when they encounter an unfamiliar object and ask another person for its name. In ALMD, the detection of novel samples is dynamic and differs from traditional out-of-distribution (OOD) detection in that the set of in-distribution (ID) classes expands as new classes are learned during application, whereas ID classes is fixed in traditional OOD detection. Learning is also different from classic supervised learning because in ALMD, we learn the encountered new classes \textit{immediately} and \textit{incrementally}. It is difficult to retrain the model from scratch using all the past data from the ID classes and the novel samples from newly discovered classes, as this would be resource- and time-consuming. 
Apart from these two challenges, ALMD faces the data scarcity issue because instances of new classes often appear sporadically in real-life applications. To address these issues, we propose a novel method, \sys, which performs dynamic OOD detection and incremental learning of new classes on the fly.
Empirical evaluations will demonstrate the effectiveness of \sys.
\end{abstract}

\end{frontmatter}


\section{Introduction}
\label{sec.intro}

{\color{black}A traditional machine learning application starts by training a model. Once the model accuracy is satisfactory, it is deployed to the application. During the application process, the model is fixed, i.e., no change to the model can be made. It assumes that the classes seen in the application must have been seen in training. This is commonly known as the \textit{closed-world assumption} \citep{fei2016breaking,kim2025open,bendale2015towards}, meaning no samples from unseen classes can appear in applications. In contrast, the real world is an open environment, full of unknowns and novelties, also known as \textit{out-of-distribution} (OOD) objects. To function effectively in this \textit{open world}, an AI agent must continuously learn on the fly after deployment, rather than relying on periodic offline retraining initiated by human engineers. This means the \textit{deployed model} should not be frozen, and more knowledge can be learned during application. We call this paradigm the \textit{Autonomous Learning after Model Deployment} (\textbf{ALMD}).} 
ALMD needs three key capabilities: 

\textbf{(1)} detecting OOD samples continually based on the current set of classes (called \textit{in-distribution} (ID) classes) that have been learned, 

\textbf{(2)} obtaining the class labels of the detected OOD samples, and 

\textbf{(3)} learning the OOD samples on the fly incrementally or continually. 
This is the \textit{class-incremental learning} (CIL) setting of continual learning (CL) as the system learns more and more classes. 

This paper focuses on (1) and (3). For (2), we do as humans do. When we humans encounter unknown objects, we usually ask others for their names (i.e., \textit{class labels}). We assume the AI agent can ask human co-workers or other agents to provide labels for the detected OOD samples.\footnote{It is possible to use a vision-language model to help assign class labels. But it cannot guarantee correctness. We leave this to our future work.} Furthermore, in ALMD, the data comes in a stream, and OOD detection and learning of the detected OOD samples are done online. {\color{black}ALMD is thus a \textbf{continual and autonomous learning} paradigm. 
\textit{\textbf{Autonomous}} means that the AI agent \textbf{takes full control} of its learning process and learns from its \textbf{own experiences} \cite{liu2023ai}:
(1) It discovers its tasks (OOD classes) to learn, (2) acquires class labels for the detected OOD samples through its interaction with human co-workers or other agents, and (3) learns the new classes incrementally. The whole process involves no human engineers. 

\textbf{ALMD Problem Setting}. Since ALMD learns continually after model deployment, the initially deployed model $M$ is assumed to be well-trained with a set of initial classes $C$ of labeled data. After $M$ is deployed in its application, it detects and learns more and more new classes. At the steady state, the set of all classes that the system has encountered is $C^A = C \cup C^L \cup C^E$, where $C^L$ is the set of new classes that have appeared after deployment and are \textit{well learned} after seeing a good number of training samples, and $C^E$ is a set of emerging new classes that have been seen but are \textit{not well learned yet}, i.e., not enough labeled training data have been seen to well-learn the classes. We denote $C^+ = C \cup C^L$ as the set of well-learned classes (ID classes) so far and $C^N = C^L \cup C^E$ as the set of all new classes seen after deploying $M$. With incremental learning of new classes, $M$ becomes $M^+$, covering all classes in $C^A$.
Each iteration of ALMD performs two main functions.

\textbf{(1).} \textit{OOD Detection and Classification}. $M^+$ detects whether each incoming test sample $x$ is OOD. If not, it is classified as one of the classes in $C^+$. OOD classes include those emerging classes in $C^E$ as they still need some more data to be well-learned, but OOD detection can leverage the already-seen samples of these classes. This is different from existing OOD detection or continual OOD detection~\citep{rios2022incdfm} (which detects OOD cases in continual learning). 

\textbf{(2).} \textit{Incremental Learning}. The system learns each detected OOD sample $x$ after obtaining its class label by asking a human or a knowledgeable agent. 
If $x$ is assigned a class in $C$, do nothing. If $x$ is assigned a class label in $C^N$, the current model $M^+$ is updated.
\vspace{2mm}

{\color{black}ALMD is thus related to three main areas of research, (1) \textbf{OOD detection}~\citep{yang2021generalized}, (2) class incremental learning (CIL) \footnote{\textit{Class-incremental learning} (CIL) is a setting of continual learning that aims to learn a sequence of tasks incrementally, where each task consists of one or more classes to be learned. The classes in the tasks are disjoint. At test time, no task-related information, e.g., task-identifier, is given.} in continual learning \citep{van2019three,de2021continual}, particularly \textbf{online continual learning} or online CIL as online CIL also learns from the streaming data \cite{aljundi2019gradient,koh2022online,mai2022online}, and (3) \textbf{open world learning}~\cite{bendale2015towards,fei2016learning,kim2025open}.}  These topics have been studied separately. However, OOD detection in ALMD has to be done continually (i.e., \textit{continual OOD} (C-OOD) \textit{detection}), unlike the traditional static OOD detection with a set of fixed \textit{in-distribution} (ID) classes. The number of ID classes in C-OOD detection increases as the AI agent learns new classes of objects. ALMD is also very different from CIL or online CIL because CIL or online CIL does not do OOD detection. It also faces the major challenge of \textit{\textbf{catastrophic forgetting}} (\textbf{CF}). CF refers to the phenomenon that the learner needs to modify the parameters learned for previous tasks in learning the new task, which may cause performance degradation for previous tasks. {\color{black}ALMD is also different from open world learning (OWL) \cite{bendale2015towards,fei2016learning}. OWL still works in the pre-deployment stage, not in the post-deployment stage as we do. They are basically offline CIL that can also do OOD detection. The ID classes in OWL are only the set of classes learned in pre-deployment, which is fixed. In our case, the set of ID classes increases as more classes become well-learned classes post-deployment. 
\cite{kim2025open} proposed a theoretical framework that is suitable for ALMD, but it does not present any algorithm. Its empirical work is only on CIL with no mechanisms for continual OOD detection. We will discuss more about the topic in Section~\ref{sec.related}. }


This paper proposes a novel approach called 
\sys~(\textit{P}ost-deployment Learning based on \textit{L}inear \textit{D}iscriminant \textit{A}nalysis) 
to learn in the ALMD setting, i.e., performing the above two main functions. The method is based on \textit{linear discriminant analysis} (LDA)~\citep{pang2005incremental}, which obtains its features from a pre-trained model (PTM). LDA assumes that given the class, the data follows a normal distribution with a mean and a covariance matrix. It further assumes that the class covariances are identical, i.e., all classes share one covariance matrix but have different means. LDA uses the means and covariance for classification. However, LDA is not suitable for OOD detection because LDA is based on the likelihood ratio, which is only suitable for closed-world classification, as OOD detection needs a measure of absolute distance from a sample to a distribution. In this work, we use Mahalanobis distance (MD) and a related method for OOD detection with a novel idea.

After obtaining the label of a detected OOD sample, \sys~learns it immediately. Each new class still uses the same shared covariance matrix learned initially in $M$, but the mean of the class is updated. The pipeline of \sys~is given in Figure~\ref{fig:openld_diagram}. It may \textbf{sound highly limiting} that LDA uses only the features from a PTM and assumes the same covariance matrix for all classes. However, as shown in Table~\ref{table:random_total} in Sec.~\ref{sec.results}, \sys~achieves a level of accuracy very close to the joint training upper bound accuracy using the pre-trained model ViT-B/14-DINOv2~\cite{oquab2023dinov2}, which has never been achieved before.

\vspace{2mm}
This paper thus makes the following contributions. 


\begin{figure*}
  \centering
  \includegraphics[width=\linewidth,height=\textheight,keepaspectratio]
  {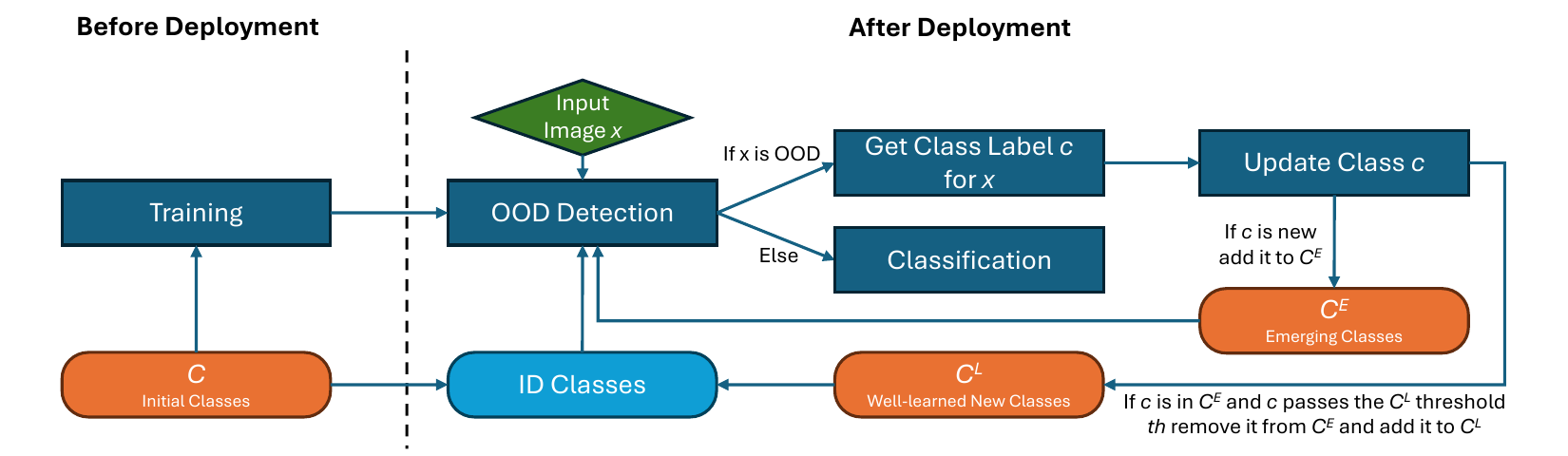}
  \vspace{0.5mm}
  \caption{Pipeline of the proposed \sys~method. ID (in-distribution) classes, which are used in OOD detection, include both $C$ and $C^L$.  }
  \label{fig:openld_diagram}
  \vspace{6mm}
\end{figure*}

\vspace{1mm}
\textbf{1.} It proposes a realistic ALMD setting, which is important as AI agents working in the real open world need to continually learn new knowledge on the fly autonomously from its own experience after deployment to make it more and more knowledgeable. 

\vspace{1mm}
\textbf{2.} It proposes a novel approach based on incremental updating of the model with a shared covariance and different means for different classes, which has \textbf{no CF} because \sys~does not do parameter updating after deployment and gives remarkably accurate results without many training samples from each new class. This is particularly important because it is hard to obtain many labeled samples after model deployment and it has not been done before. 

\vspace{1mm}
\textbf{3.} In continual OOD (C-OOD)  detection, we not only use the ID classes but also already detected OOD samples to help detect more OOD data more accurately. To our knowledge, this has not been done before either. 

\vspace{1mm}
Experiments have been conducted to demonstrate the effectiveness of the proposed \sys. The \textbf{code of \sys}~is available at \cite{PLDA_github}.  

\section{Related work}
\label{sec.related}

\textit{OOD Detection.}~OOD detection has been studied under many names, e.g., novelty or outlier detection, anomaly detection, OOD detection, and open set recognition \citep{fei2016breaking,ghassemi2022comprehensive,
bendale2015towards,
malinin2019ensemble},
~In recent years, deep learning approaches have produced state-of-the-art results \citep{yang2021generalized}. One popular category of methods uses logits to compute OOD scores~\citep{hendrycks2017baseline}.
Some other works also use additional mechanisms \citep{
sun2022dice,
liu2023gen}. Many also improve the architecture and features \citep{
huang2021mos,
sun2022dice}. Yet, some others use ensembles~\citep{lakshminarayanan2017simple}. 
Some approaches also expose the system to some OOD data during training~\citep{hendrycks2018deep,
papadopoulos2021outlier}. Some work also clusters the detected OOD samples into classes~\citep{han2021autonovel}, which we don't do as we learn each OOD sample right after it is detected. 

Our OOD detection method is most closely related to distance-based methods \citep{lee2018simple, ren2021simple}. However, our work does both OOD detection and continual learning. Unlike existing OOD detection methods, the number of ID classes in our case is not fixed but continues to increase. We also use newly identified OOD data (i.e., seen OOD classes that are not well-learned yet) to detect more OOD data.

\textit{Continual learning.} The existing work mainly focuses on overcoming CF
\citep{ke2022continualsurvey,wang2023comprehensive}.
Existing methods belong to several categories. \textit{Regularization}-based methods deal with CF in learning a new task by using a regularizer to penalize changes to parameters that are important to previous tasks  \citep{Kirkpatrick2017overcoming, 
ahn2019neurIPS}. \textit{Replay}-based methods store some data from previous tasks. When learning a new task, the saved data and the new task data are used jointly to train the new task while also adjusting the previous task parameters so that their performance will not deteriorate significantly \citep{
aljundi2019online,chaudhry2019continual}.
\textit{Pseudo-replay}-based methods build a data generator to generate previous task data to replace the replay data~\citep{
wu2018memory,
Kemker2018fearnet}.
\textit{Parameter-isolation}-based methods use masks to protect the learned models for previous tasks so that they will not be updated in learning a new task, which avoids CF \citep{Mallya2017packnet,
Serra2018overcoming,wortsman2020supermasks}. \textit{Orthogonal projection}-based methods learn each task in an orthogonal space to the previous task spaces to reduce CF \citep{zeng2019continuous,lin2022beyond}. Recently, the parameter-isolation approach and OOD detection are combined for class-incremental learning (CIL) \citep{kim2022theoretical}. However, this approach is not for open-world continual learning, but for traditional CIL.  

Most of the above methods were proposed for \textit{offline continual learning} (CL). 
Our work is more related to \textit{online Cl}, which learns from a data stream
\citep{mai2022online}. There are many existing online CL methods~\citep{
aljundi2019online,
prabhu2020gdumb,
mai2021supervised,
bang2022online,
koh2022online,mi2020generalized}. 
However, none of these methods does OOD detection, which makes it inapplicable to ALMD. 

Our work is closely related to the work in \cite{hayes2020lifelong,momeni2025continual}. \cite{hayes2020lifelong} uses \textit{incremental linear discriminant analysis} (ILDA)~\citep{pang2005incremental} for online continual learning, but it does not detect OOD instances or work in ALMD after model deployment. Further, our work does not use ILDA. {\color{black} \cite{hayes2020lifelong} already showed that even with a fixed covariance matrix, a certain amount of base classes is sufficient to create a continual learning model that is robust to increases in the number of classes and distribution changes. {\color{black}{It is discussed in more detail in Section \ref{sec.results}.}} \cite{momeni2025continual} is based on a kerneled LDA for offline CL. 
} 


\textit{Open world learning}. \cite{bendale2015towards} can incrementally learn new classes, similar to CIL. It still works in the pre-deployment stage. In post-deployment or testing, the system can do classification (known classes) and OOD detection (unknown classes). We learn in the post-deployment stage, i.e., using OOD detection to identify OOD samples from unseen classes during model application and then learn the unseen classes into the model on the fly based on each detected OOD sample.  
\cite{fei2016learning} works in a similar setting as the method in \cite{bendale2015towards}. \cite{kim2021unified} performed a theoretical study of open world continual learning, but it offers no method for OOD detection.  
\cite{gummadi2022shels} proposed SHELS for OOD detection and CL. Learning is still only in the pre-deployment stage. 
It does not integrate OOD detection and CL like ours. Their two functions can only be evaluated separately, and their CL is not on streaming data but on traditional offline CL. 
\cite{rios2022incdfm} proposed IncDFM that detects OOD using a pre-trained model. However, it does not continually learn. \cite{roy2022class} investigated class-incremental novel class discovery (class-iNCD), focusing on discovering new classes. \cite{he2022out} addressed OOD detection in an unsupervised setting. In summary, none of these methods integrates OOD detection and continual learning,  allowing the system to learn on the fly after model deployment. 

\section{Proposed approach: \sys}
\label{sec.proposed}


We now present the proposed approach \sys~for solving ALMD. 
We start with the key \textit{\textbf{challenges}} of ALMD and the main idea of the proposed techniques and their \textit{\textbf{novelties}}. Recall that ALMD has two main steps: 
(1) continual OOD (C-OOD) detection, and 
(2) class incremental learning (CIL). 
As discussed earlier, we assume that the class label for each detected OOD sample can be obtained by asking a human or another agent while working with them. 

Both steps are highly challenging. For (1), C-OOD detection is dynamic. A traditional OOD detection model is built based on a set of fixed ID classes. The key novelty of our C-OOD detection is that we also use the identified OOD samples to detect more OOD samples. For (2), the key challenge is that the AI agent should not ask human users for labels of the detected OOD samples too many times, which means we must 
have a strong learning capability without using many labeled samples. 
For both (1) and (2), there is also the challenge of \textit{catastrophic forgetting} (CF). \sys~deals with all these challenges with the help of \textit{linear discriminant analysis} (LDA). 

\subsection{Linear discriminant analysis (LDA)}

LDA is a statistical classification method that assumes each class is a normal distribution with parameters of covariance and mean for each class, i.e., $(\Sigma_i,\mu_i)$ 
\citep{fisher1936use}. 
Most LDA methods also make the simplifying homoscedasticity assumption that the class covariances are identical, i.e., $\Sigma_i = \Sigma_j = \Sigma$ for $i \neq j$. Thus, the differences between different classes are in only their means, $\mu_i$'s. This assumption is particularly useful for continual learning as the system does not have to save one covariance matrix for each class, which can consume a huge amount of memory as more classes are learned. 
LDA also makes it possible without using any replay data. 
\subsection{\sys~Method}
\label{sec.openld}

The proposed method \sys~uses a \textit{pre-trained model} (which is frozen throughout), \textit{a continual OOD detection method}, and the \textit{linear discriminant analysis} (LDA) method to solve the ALMD problem. The pre-trained model will be described in the experiment section. 
The proposed system \sys~consists of the following steps.  

\textbf{1. Building the Initial Model $M$.} \sys~uses the pre-trained model $f$ to provide the features for input samples, which are used by LDA to build a classifier $M$ using the initial classes $C$. As mentioned earlier, LDA's classifier building is based on a mean $\mu_i$ for each class $i \in C$ and a single shared covariance matrix $\Sigma$ across all classes. Thus, it produces the shared $\Sigma$ and a separate mean $\mu_i$ for each class $i \in C$. 
The resulting model $M$ is deployed in its application (Figure~\ref{fig:openld_diagram}). 

\textbf{2. Post-deployment Continual Learning.} After deployment, it continues to learn, which will update $M$ after new classes are incrementally learned and $M$ becomes $M^+$. In the continual learning process, $\Sigma$ remains unchanged or frozen and it is also used by the newly detected classes, which, as discussed in Section~\ref{sec.intro}, is by no means limiting.
Each iteration has two sub-steps.

\vspace{+1mm}
\textbf{2.1. Continual OOD Detection and Classification.} \sys~uses $M^+$ to detect whether each sample $x$ in the online stream is an OOD sample. If not, it is classified to its class (see Figure~\ref{fig:openld_diagram}). Note that $M^+$ is $M$ initially. \sys~employs the covariance matrix $\Sigma$ and the $\mu_i$'s for all classes encountered or seen so far to perform the tasks. 

At a steady state, the set of all classes that the system has encountered is $C^A = C \cup C^L \cup C^E$, where $C$ is the initial set of classes learned in $M$, $C^L$ is a set of \textit{well learned} new classes after seeing a good number of instances, and $C^E$ is a set of emerging classes that have been seen but are \textit{not yet well learned}. 
Note that, in the ALMD setting, the classes are updated multiple times, with one sample at a time, throughout the AI agent's lifetime, as it encounters a sample from an OOD class.
\textbf{Well learned} means that the mean of the class 
does not change much after more samples are added. A class becomes well-learned if its mean is updated at least $th$ times, where $th$ is the selected convergence threshold. 
We denote $C^+ = C \cup C^L$ as the set of well-learned classes so far and $C^N = C^L \cup C^E$ as the set of all new classes seen after the deployment of $M$. With incremental learning, $M$ becomes $M^+$, covering all classes in $C^A$.

What is important here is that OOD detection not only uses the classes in $C^+$ but also leverages the covariance $\Sigma$ and the current un-converged means of the classes in $C^E$ to detect OOD samples belonging to $C^E$ and other new classes. To our knowledge, no existing method does that. This is advantageous because a new sample may be similar to a class in $C^E$, which makes OOD detection more effective. In Sec.~\ref{sec.ood-detction}, we discuss the OOD detection methods used in our \sys. If a test sample $x$ is near a class in $C^E$, it is also considered an OOD sample.

\textbf{2.2 Continual Learning - \textit{class-incremental learning} (CIL).} Here the continual learning setting is CIL, which incrementally learns more and more new classes. 
Specifically, \sys~learns each detected novel instance $x$ after obtaining its class label by asking a human user or another knowledgeable agent. If $x$ is assigned a class label in $C$, do nothing (i.e., no learning). If $x$ is assigned a class label $c_i$ in $C^N$, the current model $M^+$ is updated by updating the mean $\mu_i$ of the class $c_i$ (covariance matrix $\Sigma$ is not changed) as  
\begin{align}
    \mu_i \leftarrow \frac{n_i\mu_i + z}{n_i + 1}, 
    \label{eq.mu}
\end{align}
where $z$ is the feature $f(x)$ obtained from the pre-training model $f$ and $n_i$ is the number samples seen so far in class $i$. 

This approach has two desirable properties. 

(1) \sys~has no catastrophic forgetting (CF) during ALMD as we use a frozen pre-trained model (or feature extractor),\footnote{By no means that using a pre-trained model without feature learning is a weakness. As we will see in the experiment section, the constantly advancing \textit{pre-trained models} produce rich features for CL. The baselines that learn features produce poorer results even with replay data.} a fixed and shared covariance $\Sigma$, and a running mean for each class, which is independent of those of other classes. Thus, there is no interference across classes. 

(2) Again, due to the sharing of covariance matrix $\Sigma$ by all old and new classes, we achieve strong learning results with a small number of examples because, for each detected new class, \sys~only updates its mean based on the identified samples of the class. 

\subsection{OOD Detection Methods}
\label{sec.ood-detction}

Since \sys~produces a shared covariance $\Sigma$ and one mean $\mu_i$ for each class $i$, we can naturally use $\Sigma$ and $\mu_i$ related OOD detection methods, i.e., \textit{Mahalanobis distance} (MD) and relative \textit{Mahalanobis distance} (RMD). Each of these methods produces a confidence score using all classes $k \in C^A$ for the given feature vector $z = f(x)$, where $x$ is the input. If the confidence score is below a \textit{threshold} level, or it belongs to any class in $C^E$, that input is marked as OOD. Note that apart from these methods, there are numerous existing OOD detection methods (see Sec.~\ref{sec.related}). However, since our approach does not train a neural network, most existing methods are not suitable for use in \sys. This is due to a few reasons. First, the number of our ID classes is not constant but keeps increasing, which means that the OOD detection model needs to be updated, causing CF. Second, since we cannot train all classes together in CL, those OOD detection methods that need to use logits are not applicable. Third, methods based on sample distances, e.g., KNN, are also inapplicable as we cannot save the past data in CL.

\subsubsection{Mahalanobis distance (MD)}

Mahalanobis distance~\citep{mclachlan1999mahalanobis} measures the distance between a data point (a feature vector in our case) and a normal distribution using the class mean vector $\mu$ and the covariance $\Sigma$, which is suitable for OOD detection~\cite{lee2018simple}. Note that, each class mean $\mu_i$ and covariance $\Sigma$ for the data used in building the initial model $M$ are estimated as: $\mu_i= \frac{1}{N_i} \sum_{k:y_{k}=i} z_k$ and $\Sigma = \frac{1}{N} \sum_{i \in C} \sum_{k:y_k=i} (z_k - \mu_i) (z_k - \mu_i)^T $, where $N$ denotes the number of samples, $N_i$ denotes the number of samples of class $i$, and $z_k$ is the feature of input sample $x_k$ obtained from the pre-trained model, i.e., $z_k=f(x_k)$. $\Sigma$ is the same for new classes, while 
$\mu_i$ for the new classes are incrementally computed using Eq.~\ref{eq.mu}. 

For $z =f(x)$ of a test sample $x$, we compute MD as, 
\begin{align}
    MD_i(z;\mu_i,\Sigma) = \sqrt{(z - \mu_i)^T \Sigma^{-1} (z - \mu_i)}
\end{align}
where $\Sigma^{-1}$ is the inverse of covariance matrix. The confidence score $c$ is described as,
\begin{align}
    c(z) = \mathop{max}_{i \in C^+} \{1/ MD_i(z;\mu_i,\Sigma)\}
\end{align}

\subsubsection{Relative Mahalanobis distance (RMD)}

As noted in \citep{ren2021simple}, MD has some limitations regarding the detection of OOD data and they proposed RMD by applying a simple addition to MD. It computes an additional mean $\mu_C = \frac{1}{N} \sum_{k = 1}^N z_k$ and covariance  $\Sigma_C = \frac{1}{N} \sum_{k = 1}^N (z_k - \mu_C) (z_k - \mu_C)^T $, which, in our case, are only calculated based on the initial data with $C$ classes used in building model $M$. RMD is computed as, 

\begin{equation}
\begin{aligned}
&RMD_i(z;\mu_i,\Sigma,\mu_C,\Sigma_C) = \\
        &MD_i(z;\mu_i,\Sigma) - MD_A(z;\mu_C,\Sigma_C),
\end{aligned}
\end{equation}

where $MD_A(z;\mu_C,\Sigma_C) = \sqrt{(z - \mu_C)^T \Sigma_C^{-1} (z - \mu_C)}$. The confidence score is~\citep{ren2021simple}
\begin{align}
    c(z) = \mathop{max}_{i \in C^+} \{ - RMD_i(z;\mu_i,\Sigma,\mu_C,\Sigma_C)\}
\end{align}



\section{Experimental evaluation}
\label{sec.experiments}
We now evaluate the proposed method \sys. We will see that \sys~can produce accuracy very close to those from joint fine-tuning, which learns all classes together as a single task in many epochs to reach the best classification accuracy. It is considered the upper bound of continual learning. 

\subsection{Datasets, compared methods, pre-trained model, and implementation}
\paragraph{Datasets.} We use three benchmark image classification datasets in our experiments. 

\textbf{(1) CIFAR-10} \citep{krizhevsky2009learning}: It contains 60,000 images, 
50,000 training images and 10,000 testing images, distributed evenly across 10 classes.

\textbf{(2) CIFAR-100} \citep{krizhevsky2009learning}: It contains 50,000 training images (500 per class) and 10,000 testing images (100 per class) of 100 classes. 

\textbf{(3) TinyImageNet} \citep{le2015tiny}: It contains 200 classes, each with 500 training images. The validation set includes 50 images per class. Since the test data labels are unavailable, we use the validation set for testing.

\textbf{-- ID Class Set} and \textbf{OOD Class Set}: For each experiment dataset, we divide the classes in the dataset into an equal number of \textit{ID} (\textit{in-distribution}) classes and \textit{OOD} (\textit{out-of-distribution}) classes. The \textbf{ID class set} is used to build the initial model $M$ for deployment, while the \textbf{OOD class set} is used in incremental learning after model deployment. 

\textbf{-- ID+OOD APP Set}: We further divide the training set of each class in the ID class set into \textit{\textbf{ID Train set}} and \textit{\textbf{ID APP set}}. We do the same for the OOD class set. \textit{\textbf{ID+OOD APP set}} includes the data from both the ID and OOD APP sets. The ID+OOD APP set simulates the application (APP) data from a real-life data stream that needs to be classified.  

For CIFAR-100 and TinyImageNet, each class has 500 samples in the original training set. After the split, the ID Train set has 450 samples and the ID+OOD APP set has 50 samples per class. CIFAR-10 has 5000 samples per class in its original training set, but to simulate the situation where the system does not ask the human user too many questions, we selected 4500 samples per class for the ID Train set and 50 samples per class for the ID+OOD APP set. We use a small number of samples in the ID+OOD APP set for each class to simulate the situation where in the application or after deployment, we don't see so many OOD samples. The rest of the OOD class data is not used in our experiment.

\textbf{-- Pre- and Post-Deployment:} In pre-deployment, we perform joint training using LDA and a pre-trained model to build the model $M$ using only the ID Train set. ID+OOD APP set is used only post-deployment. 


\vspace{1mm}
\noindent
\textbf{Compared Methods.} Although there are several related papers~\citep{bendale2015towards,gummadi2022shels,rios2022incdfm,roy2022class}, as discussed in Sec~\ref{sec.related}, no existing system can perform ALMD after model deployment as proposed in this paper. 

{\color{black}
Since our setting is closely related to online CL, we compare our method with 7 online CL baselines:

--\textbf{LwF} \cite{Li2016LwF}: A regularization-based method that uses knowledge distillation to preserve the performance on previous tasks to deal with CF without storing old data.

--\textbf{iCaRL} \cite{rebuffi2017icarl}: A replay-based method that maintains a small exemplar memory and uses a nearest-mean-of-exemplars classifier.

--\textbf{AGEM} \cite{chaudhry2019efficient}: It mitigates forgetting by projecting gradients to avoid interference with stored memory samples.

--\textbf{ER} \cite{chaudhry2019continual}: A replay-based method that replays a small buffer of past examples with random sampling alongside new data during training.

--\textbf{MIR} \cite{aljundi2019online}: A replay-based method that prioritizes memory samples most vulnerable to forgetting for retrieval.

--\textbf{GDumb} \cite{prabhu2020gdumb}: A replay-based method that stores a balanced memory and retrains from scratch for evaluation.

--\textbf{GACL} \cite{zhuang2024gacl}: A recent method that uses an analytic solution to avoid forgetting by achieving a weight-invariant property.
}


To ensure a fair comparison, we replaced their backbone model with ViT-B/14-DINOv2, and added adapters to prevent CF \cite{kim2022theoretical}. Note that these methods do not do OOD detection. 

For each baseline, ID Train set is used to learn ID classes to build the initial model $M$ as the first task, and OOD APP set is used to learn the OOD classes incrementally. However, since these baselines don't detect OOD samples, we assume that the baseline methods can do perfect OOD detection (thus ID APP set is not used). Even in this ideal case, these baseline methods perform poorer than \sys.

\textbf{Two upper-bound methods} are also created, which are not continual learning methods.

-- \textbf{Joint LDA:} This method applies LDA to learn a classifier using the data from all classes jointly, i.e., ID Train set and OOD APP set. ID APP set is not used as there is no post-deployment learning or OOD detection in this setting. 
This method gives the \textbf{upper bound results} of LDA. 

\textbf{-- Joint Fine-tune:} This method fine-tunes the pre-trained model using ID Train set and OOD APP set. We used the AdamW optimizer with a learning rate of 0.0001, CosineAnnealingLR scheduler, batch size of 128, and trained the model for 30 epochs, which is sufficient for convergence.

\textbf{Ablations:} We also create variations of the proposed method \sys~for ablation experiments.

\textbf{--~\sys[X]:} This creates three variations of \sys, using different OOD detection methods (X), i.e., MD or RMD for OOD detection (see Section~\ref{sec.ood-detction}). 

\textbf{--~\sys[X](-$C^E$):} This also creates two variations of \sys~\textbf{\textit{without}} leveraging the un-converged means of the classes in $C^E$ to help detect OOD samples. These 
will show that using $C^E$ in OOD detection is helpful. 



\textbf{Pre-trained Models.} For feature extraction, our primary pre-trained model is \textbf{ViT-B/14-DINOv2}~\cite{oquab2023dinov2}, which is a self-supervised model with no class information leak. As another model we used, \textbf{DeiT-S/16-Kim} \citep{kim2022multi}, based on DeiT-S/16 \citep{touvron2021training}, was pre-trained using the labeled ImageNet-1k data. To prevent class information leaks between the pre-training and continual learning phases, 389 classes in ImageNet-1k similar to classes in CIFAR-10, CIFAR-100, and TinyImageNet, were excluded and pre-training was performed with the remaining 611 classes to produce DeiT-S/16-Kim. We also use another self-supervised model, \textbf{ViT-B/16-DINO} \citep{caron2021emerging}.

\textbf{Implementation and Resource Usage.} 
We used the LDA implementation in~\citep{hayes2020lifelong} and the \textbf{ViT-B/14-DINOv2}~\cite{oquab2023dinov2} pre-trained model. 
We run on a machine with an AMD EPYC 7502 32-Core Processor and NVIDIA RTX A6000 GPU. Each experiment requires approximately 6 GB of GPU memory and takes an average of 10 minutes. 

\textbf{OOD Detection and $C^L$ Thresholds.} In each setup, confidence scores for OOD detection are computed using two alternative methods, RMD and MD, and an image input is considered OOD if its confidence score is less than a certain threshold. These thresholds are set to 0.012 and 4.9 for RMD and MD, respectively. They are chosen empirically to ensure that the precision and recall for the OOD data are similar for each method. The threshold for $C^L$ is empirically set to 30, which we will study later. 


\begin{table*}[!hbt]
  \caption{
  Performance comparison of \sys~with baselines and ablation of \sys~using two different OOD detection methods (MD and RMD) on CIFAR-10, CIFAR-100, and TinyImageNet datasets in the \textit{Random ID+OOD APP Data Arrival} setup (Setup 1). F-score gives the OOD detection performance. Note that baselines do not have F-score values as we assume their OOD detection is perfect (ideal case). Buffer size is the replay buffer size. 
  {\color{black} $C^L$ threshold is set to 30 for \sys, higher values result in even higher accuracies and F-scores for the proposed method, with a cost of more number of asks.}
  }
   \vspace{2mm}
  \label{table:random_total}
  \scriptsize
  \centering
  \resizebox{\linewidth}{!}{
      \begin{tabular}{lccccccccc} \\
        \toprule
         & &
         \multicolumn{2}{c}{\bf CIFAR-10}  & 
         \multicolumn{2}{c}{\bf CIFAR-100} & 
         \multicolumn{2}{c}{\bf TinyImageNet} & 
         \multicolumn{2}{c}{\bf Average} \\ 
         \hline
        
        \bf Methods & \bf Buffer Size &
        \bf F-score  & \bf Accuracy &
        \bf F-score  & \bf Accuracy &
        \bf F-score  & \bf Accuracy &
        \bf F-score  & \bf Accuracy \\
        
        \hline
        \textit{Joint LDA} & &  & \textit{97.03{\tiny$\pm$0.00}} &  & \textit{85.73{\tiny$\pm$0.00}} &  & \textit{81.89{\tiny$\pm$0.00}} &  & \textit{88.22} \\
        \textit{Joint Fine-tune} & &  & \textit{93.22{\tiny$\pm$0.16}} &  & \textit{89.74{\tiny$\pm$0.13}} &  & \textit{85.90{\tiny$\pm$0.10}} &  & \textit{89.62} \\
        \hline
        LwF & 0 &  & 63.68{\tiny$\pm$4.57} &  & 56.77{\tiny$\pm$0.13} &  & 57.41{\tiny$\pm$0.69} &  & 59.29 \\
        AGEM & 5000 &  & 90.98{\tiny$\pm$1.07} &  & 72.48{\tiny$\pm$0.43} &  & 75.50{\tiny$\pm$0.62} &  & 79.65 \\
        GACL & 0 &  & 84.95{\tiny$\pm$0.20} &  & 79.44{\tiny$\pm$0.08} &  & 78.01{\tiny$\pm$0.13} &  & 80.80 \\
        GDumb & 1000 &  & 94.59{\tiny$\pm$0.81} &  & 81.34{\tiny$\pm$3.04} &  & 73.56{\tiny$\pm$0.03} &  & 83.16 \\
        ER & 5000 &  & 94.70{\tiny$\pm$0.52} &  & 80.44{\tiny$\pm$0.67} &  & 79.94{\tiny$\pm$0.59} &  & 85.03 \\
        iCaRL & 5000 &  & 95.48{\tiny$\pm$0.73} &  & 81.06{\tiny$\pm$0.09} &  & 79.23{\tiny$\pm$0.59} &  & 85.26 \\
        MIR & 5000 &  & 95.59{\tiny$\pm$0.39} &  & 81.19{\tiny$\pm$0.29} &  & 80.64{\tiny$\pm$0.69} &  & 85.81 \\
        
        \hline

        \sys[MD]-$C^E$ & 0 & 75.20{\tiny$\pm$0.81} & 96.61{\tiny$\pm$0.19} & 73.88{\tiny$\pm$0.15} & 83.96{\tiny$\pm$0.06} & 76.86{\tiny$\pm$0.07} & 81.68{\tiny$\pm$0.11} & 75.31 & 87.42 \\
        \sys[RMD]-$C^E$ & 0 & 75.50{\tiny$\pm$0.18} & 96.53{\tiny$\pm$0.10} & 74.48{\tiny$\pm$0.14} & 84.21{\tiny$\pm$0.09} & 78.17{\tiny$\pm$0.22} & 81.49{\tiny$\pm$0.08} & 76.05 & 87.41 \\
        \sys[MD] & 0 & 75.45{\tiny$\pm$0.70} & 96.48{\tiny$\pm$0.27} & 79.55{\tiny$\pm$0.49} & 85.12{\tiny$\pm$0.03} & 78.31{\tiny$\pm$0.14} & 81.77{\tiny$\pm$0.05} & 77.77 & 87.79 \\
        \sys[RMD] & 0 & 75.69{\tiny$\pm$0.24} & 96.74{\tiny$\pm$0.24} & 79.69{\tiny$\pm$0.02} & 85.28{\tiny$\pm$0.05} & 80.63{\tiny$\pm$0.24} & 81.70{\tiny$\pm$0.28} & 78.67 & 87.91 \\
        \hline
      \end{tabular}
  }
\end{table*}

\begin{table*}[t]
  \caption{Performance comparison of \sys~using different pre-trained models. All experiments use RMD as the OOD score method. \textbf{Joint} is the accuracy produced by joint fine-tuning the corresponding pre-trained model, considered the upper bound accuracy. Note that on CIFAR-10, due to highly imbalanced data, its Joint fine-tuning accuracy is low, but its Joint LDA accuracy is high (see Table~\ref{table:random_total}).
  } 
  \label{table:random_vits}
  \vspace{2mm}
  \centering
  \resizebox{\linewidth}{!}{
      \begin{tabular}{lcccccccccccc} \\
        \toprule
         & \multicolumn{3}{c}{\bf CIFAR-10} & \multicolumn{3}{c}{\bf CIFAR-100} & \multicolumn{3}{c}{\bf TinyImageNet} &
         \multicolumn{3}{c}{\bf Average}
         \\ \hline 
        
        \bf Pre-trained Models & 
        \bf F-score   & \bf Accuracy & \bf Joint &
        \bf F-score  & \bf Accuracy & \bf Joint &
        \bf F-score  & \bf Accuracy & \bf Joint &
        \bf F-score  & \bf Accuracy & \bf Joint \\
        \hline

        DeiT-S/16-Kim & 
         67.98{\tiny$\pm$2.32} & 81.88{\tiny$\pm$0.69} & 80.59{\tiny$\pm$1.28} & 
         65.89{\tiny$\pm$0.49} & 62.55{\tiny$\pm$0.18} & 72.35{\tiny$\pm$0.43} & 
         65.41{\tiny$\pm$0.10} & 55.72{\tiny$\pm$0.18} & 61.92{\tiny$\pm$0.33} & 
         66.42 & 66.71 & 71.62 \\

        ViT-B/16-DINO & 
        68.64{\tiny$\pm$1.37} & 89.01{\tiny$\pm$0.88} & 88.13{\tiny$\pm$0.52} & 
        70.23{\tiny$\pm$0.92} & 71.99{\tiny$\pm$0.33} & 81.72{\tiny$\pm$0.04} &
        72.86{\tiny$\pm$0.26} & 72.39{\tiny$\pm$0.19} & 81.21{\tiny$\pm$0.03} & 
        70.57 & 77.80 & 83.69 \\

        ViT-B/14-DINOv2 & 
        75.69{\tiny$\pm$0.24} & 96.74{\tiny$\pm$0.24} & 93.22{\tiny$\pm$0.16} & 
        79.69{\tiny$\pm$0.02} & 85.28{\tiny$\pm$0.05} & 89.74{\tiny$\pm$0.13} & 
        80.63{\tiny$\pm$0.24} & 81.70{\tiny$\pm$0.28} & 85.90{\tiny$\pm$0.10} & 
        78.67 & 87.91 & 89.62 \\

        \hline

      \end{tabular}
  }
\end{table*}

\begin{figure*}
  \centering
  \includegraphics[width=\linewidth,height=\textheight,keepaspectratio]{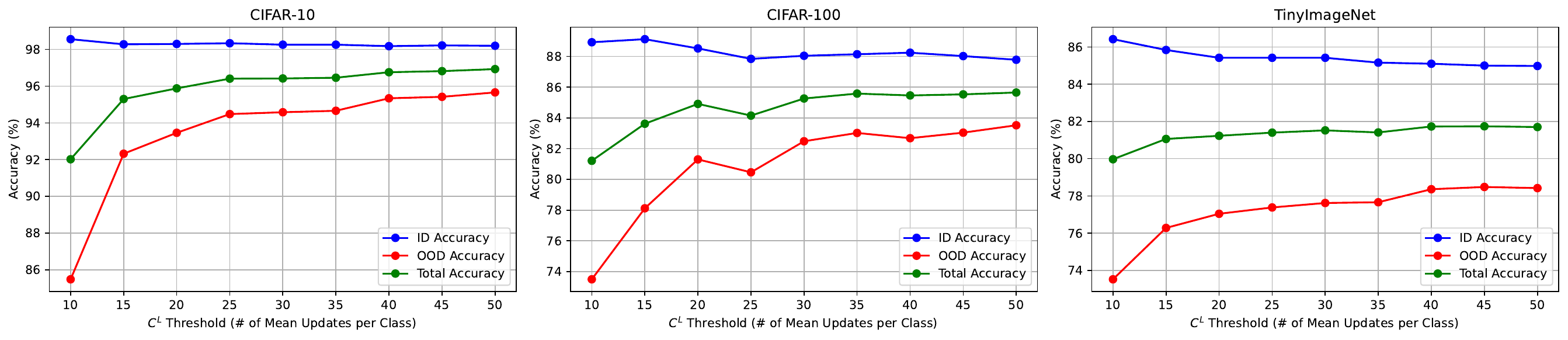}
  \caption{Accuracies of \sys~for different $C^L$ threshold values for different datasets. While there is no CF, the ID accuracy drops as expected as more classes are learned, which makes the classification harder. The decrease in the ID accuracy is minimal relative to the significant gains in the OOD accuracy, demonstrating the model’s resilience. 
  }
  \label{fig:cl_threshold}
  \vspace{2mm}
\end{figure*}

\subsection{Two experiment setups and evaluations}
\label{sec.exp-setups}

\textbf{Setup 1) Random ID+OOD APP Data Arrival.} This is \textit{\textbf{our main setup}} as it reflects real-life application scenarios. In this setup, after model deployment, the samples in the ID+OOD APP set arrive randomly in a data stream. The current model $M^+$ classifies each to its class or detects it as OOD. For each detected OOD sample (which may be correct or wrong), the proposed system \sys~asks the human user for its class label, and then \sys~incremental/continual learning is performed by updating its class mean. Clearly, in our experiment, \textit{\textbf{no human user}} is involved. The system just uses the class label of the sample in the original data. 

\textbf{Setup 2) Class-Incremental OOD APP Data Arrival.} This setup is for scenarios where an AI agent keeps going to new environments. This scenario may be rare. We use it to show the robustness of our approach. Each environment has a set of new OOD classes as a new task. This is similar to class incremental learning (CIL). In our case, we divide the classes in the OOD class set evenly into 5 tasks (5 environments). Each task contains one-fifth of random samples from the ID APP set and all samples in the OOD APP set of the classes in the task. 
The OOD samples from one task finish before the data from the next task arrives. Acquiring labels of OOD samples and updating the class means for continual learning are the same as above. 

\subsection{Results: Comparison with baselines and ablations}
\label{sec.results}




\begin{table*}[hbt]
  \caption{Performance comparison of \sys~with baselines and ablation of \sys~using two different OOD detection methods (MD and RMD) on CIFAR-10, CIFAR-100, and TinyImageNet datasets in the \textit{Class-Incremental OOD APP Data Arrival} setup (Setup 2). F-score gives the OOD detection performance. Baselines do not have F-score values as we assume their OOD detection is perfect. Buffer size is the replay buffer size.}
  
  \vspace{2mm}
  \label{table:cil_total}
  \scriptsize
  \centering
  \resizebox{\linewidth}{!}{
      \begin{tabular}{lccccccccc} \\
        \toprule
         & &
         \multicolumn{2}{c}{\bf CIFAR-10}  & 
         \multicolumn{2}{c}{\bf CIFAR-100} & 
         \multicolumn{2}{c}{\bf TinyImageNet} & 
         \multicolumn{2}{c}{\bf Average} \\ 
         \hline
        
        \bf Method & \bf Buffer Size &
        \bf F-score  & \bf Accuracy &
        \bf F-score  & \bf Accuracy &
        \bf F-score  & \bf Accuracy &
        \bf F-score  & \bf Accuracy \\
        
        \hline
        LwF & 0  &  & 18.86{\tiny$\pm$5.88} &  & 26.69{\tiny$\pm$2.67} &  & 29.45{\tiny$\pm$1.77} &  & 25.00 \\
        AGEM & 5000 &  & 80.41{\tiny$\pm$1.69} &  & 68.32{\tiny$\pm$0.62} &  & 73.83{\tiny$\pm$0.54} &  & 74.18 \\
        ER & 5000 &  & 83.36{\tiny$\pm$2.47} &  & 69.53{\tiny$\pm$1.75} &  & 73.73{\tiny$\pm$0.55} &  & 75.54 \\
        MIR & 5000 &  & 87.91{\tiny$\pm$2.96} &  & 71.16{\tiny$\pm$0.62} &  & 74.56{\tiny$\pm$1.27} &  & 77.88 \\
        GACL & 0 &  & 84.82{\tiny$\pm$0.13} &  & 79.27{\tiny$\pm$0.11} &  & 77.96{\tiny$\pm$0.15} &  & 80.68 \\
        iCaRL & 5000 &  & 91.74{\tiny$\pm$1.28} &  & 78.38{\tiny$\pm$0.17} &  & 74.70{\tiny$\pm$0.27} &  & 81.61 \\
        GDumb & 1000 &  & 92.47{\tiny$\pm$2.34} &  & 81.45{\tiny$\pm$2.23} &  & 71.76{\tiny$\pm$0.41} &  & 81.89 \\
        \hline
        \sys[MD]-$C^E$ & 0 & 74.76{\tiny$\pm$0.47} & 96.40{\tiny$\pm$0.14} & 72.79{\tiny$\pm$0.10} & 83.72{\tiny$\pm$0.11} & 75.88{\tiny$\pm$0.07} & 81.47{\tiny$\pm$0.22} & 74.47 & 87.20 \\
        \sys[RMD]-$C^E$ & 0 & 76.31{\tiny$\pm$0.92} & 96.71{\tiny$\pm$0.12} & 73.77{\tiny$\pm$0.17} & 84.33{\tiny$\pm$0.18} & 76.70{\tiny$\pm$0.09} & 80.91{\tiny$\pm$0.08} & 75.59 & 87.32 \\
        \sys[MD] & 0 & 75.32{\tiny$\pm$0.24} & 96.53{\tiny$\pm$0.05} & 78.12{\tiny$\pm$0.41} & 84.97{\tiny$\pm$0.12} & 77.64{\tiny$\pm$0.09} & 81.53{\tiny$\pm$0.10} & 77.02 & 87.68 \\
        \sys[RMD] & 0 & 76.40{\tiny$\pm$0.72} & 96.64{\tiny$\pm$0.26} & 77.83{\tiny$\pm$0.31} & 84.75{\tiny$\pm$0.29} & 79.09{\tiny$\pm$0.35} & 80.92{\tiny$\pm$0.07} & 77.77 & 87.43 \\
        \hline
      \end{tabular}
  }
\vspace{2mm}
\end{table*}

\textbf{Setup 1: Random ID+OOD APP Data Arrival:} 

Table \ref{table:random_total} shows the OOD detection F-score, and accuracy of the combined ID and OOD classes after all data in the ID+OOD APP set are seen. \sys~variants (which use $C^E$) give significantly higher F-scores than their corresponding variants without using $C^E$. The final accuracy is also better. Note that the accuracy improvement is not large as the test results are obtained after the system sees all data, at which time the means of the OOD classes have mostly converged. We also observe that the baselines, which also do feature learning, are markedly poorer even in their ideal situation, i.e., OOD detection is perfect with no errors. The most recent online CL method GACL 
does poorly. Note that Joint fine-tuning is poor for CIFAR-10 due to highly imbalanced class distribution. 0 in the Buffer Size column means that the system does not store any replay data. 

-- \textbf{\textit{{Results of Different Pre-Trained Models}.}}~Table~\ref{table:random_vits} shows that with the bigger model (DINOv2), the accuracy gets closer to the Joint fine-tuning upper bound. The OOD detection F-score also improves significantly. 
Our method performed even better than offline Joint fine-tuning for CIFAR-10, where the class imbalance is very high (4500 samples in each ID class in pre-deployment, but only 50 samples per class in each OOD class in post-deployment). 

With bigger and more powerful pre-trained models appearing constantly, the results will improve further. There will be no need to learn new features or fine-tune the pre-trained model, which causes CF. 

-- \textbf{\textit{Efficiency}.} Since \sys~uses the statistical methods LDA in learning and MD for OOD detection, which only needs to incrementally update the mean for each class and the shared covariance matrix, it is much more efficient than training in deep learning. In the post-deployment stage, \sys~performs only OOD detection and updating of the mean of each new class (no parameter training), which takes almost no time (less than 15 milliseconds), because the features are already learned in the pre-trained model. This makes \sys~especially suitable for ALMD in real time. {\color{black} Figure \ref{fig:runtime_comparison} shows the training and inference run time comparison between baselines and \sys.}

\begin{figure}
  \centering
  \includegraphics[width=\linewidth,height=\textheight,keepaspectratio]{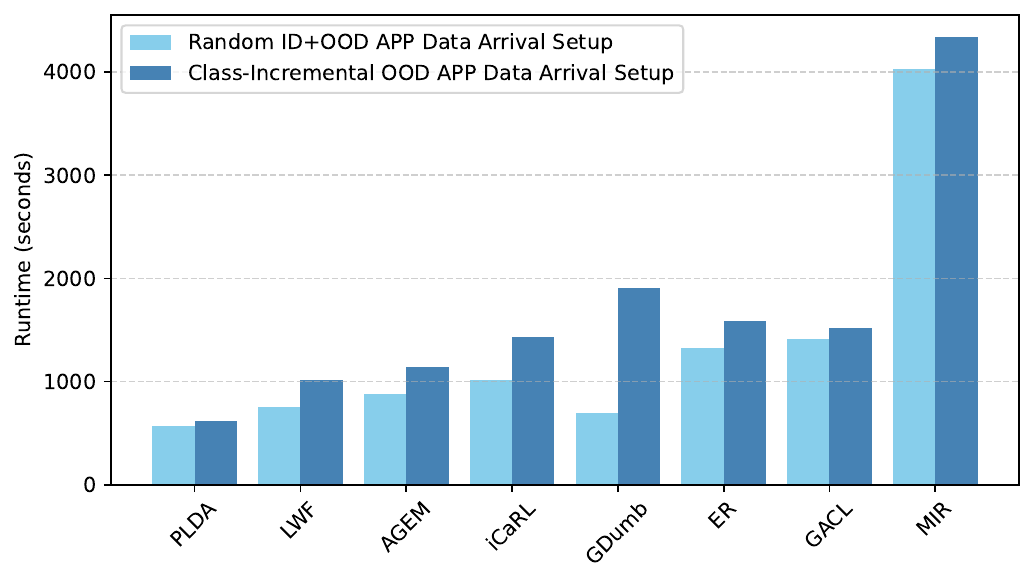}
\caption{Run time comparison. \sys~achieves the best performance and the fastest run time among all methods across both setups.}
  \label{fig:runtime_comparison}
   \vspace{7mm}
\end{figure}

-- \textbf{$C^L$ \textit{Threshold}.} The threshold \textit{th} for $C^L$ is set to 30 in the main experiments. We tested values between 10 and 50, and the results are shown in Figure \ref{fig:cl_threshold}. As the threshold (the number of mean updates or asks per class) increases, the total and the OOD class accuracy improves, but the ID class accuracy slightly drops (because more classes are considered in the classification). We chose 30 for all datasets as the threshold in our experiments because when the threshold reaches 30, the total accuracy stabilizes. It also achieves a good balance between accuracy and the number of asks (or mean updates). For example, by selecting 30 instead of 50 as the $C^L$ threshold, we can reduce the number of asks by about 26.28\%, losing only 0.36\% in the total accuracy on average over the three datasets. The total accuracy covers both the ID classes and the OOD classes. 

{\color{black}
-- \textbf{Shared Covariance Matrix.} As another ablation experiment, we explored the performance of using a separate covariance matrix for each OOD class, without sharing the covariance across classes. When tested on the CIFAR-100 dataset, we saw a drop in accuracy, from 85.28\% to 82.90\%. We hypothesize that this decrease in performance is due to the lack of sufficient samples for out-of-distribution (OOD) classes, making it challenging to construct accurate and robust covariance matrices for those classes. Our fixed shared covariance approach, on the other hand, works better because it combines information from a large number of training samples of multiple classes in the pre-deployment training, helping create a more reliable covariance estimate. This finding is supported by \cite{hayes2020lifelong}, which shows a robust performance with a fixed covariance computed from the ImageNet data. This further emphasizes the advantages of using fixed shared covariance, especially when dealing with class imbalance or limited data in the real world.

}

\textbf{Setup 2: Class-Incremental OOD APP Data Arrival:}
Table \ref{table:cil_total} shows the final classification accuracy and OOD detection F-score. \sys~shows a very similar performance to Setup 1, which shows the robustness of \sys~. The baselines are poorer because the tasks are incrementally arriving which causes more CF. In Setup 1, data from different classes arrive randomly, which is like arriving together with only one epoch of training. 



\section{Conclusion}
This paper proposed the setting of \textit{Learning After model Deployment} (ALMD). {\color{black}It enables the AI agent to learn based on its own experience in an autonomous manner. Although similar ideas like open world learning have been around for some time and some preliminary work has also been done, none of them have truly implemented learning after model deployment so that the AI agent can learn to classify more and more classes on the fly while working. We believe that ALMD is getting more and more important as more and more AI agents are deployed in real-life applications. It is highly desirable that these agents can learn continually after deployment by themselves to become more and more knowledgeable over time.} This paper also proposed a method (called \sys) based on LDA and a pre-trained model with several novel techniques to improve OOD detection in the ALMD process and to learn new classes easily by only updating class means, which has no catastrophic forgetting (CF) introduced in traditional continual learning due to network parameter updating in learning new classes or tasks. Experiment results have demonstrated the effectiveness of the proposed method \sys.  

One limitation of our work is that our OOD detection method may be effective, but it may not be state-of-the-art (SOTA). As explained in Sec.~\ref{sec.ood-detction}, it is difficult to use a current SOTA OOD detection method because they are unsuitable for continual OOD detection as they can cause serious catastrophic forgetting (CF) in our continual OOD detection setting. More work is needed to design more effective OOD detection methods for the continual learning context.   

\section*{Ethics Statement}
{\color{black}Our experiments used public domain benchmark datasets for image classification, which have been widely used in continual learning evaluations. They do not contain any unethical or offensive images. Our proposed system constructs a statistical model to categorize images into distinct classes. It does not generate any unethical content.
} 

\begin{ack}
The work of Derda Kaymak and Bing Liu was supported in part by three NSF grants (IIS-2229876, IIS-1910424, and CNS-2225427), and an NVIDIA's Academia Grant, which provides cloud compute via its Saturn Cloud. 
\end{ack}
\bibliography{mybibfile}

\begin{thebibliography}{62}
\providecommand{\natexlab}[1]{#1}
\providecommand{\url}[1]{\texttt{#1}}
\expandafter\ifx\csname urlstyle\endcsname\relax
  \providecommand{\doi}[1]{doi: #1}\else
  \providecommand{\doi}{doi: \begingroup \urlstyle{rm}\Url}\fi

\bibitem[Ahn et~al.(2019)Ahn, Cha, Lee, and Moon]{ahn2019neurIPS}
H.~Ahn, S.~Cha, D.~Lee, and T.~Moon.
\newblock Uncertainty-based continual learning with adaptive regularization.
\newblock In \emph{NeurIPS}, 2019.

\bibitem[Aljundi et~al.(2019{\natexlab{a}})Aljundi, Caccia, Belilovsky, Caccia,
  Lin, Charlin, and Tuytelaars]{aljundi2019online}
R.~Aljundi, L.~Caccia, E.~Belilovsky, M.~Caccia, M.~Lin, L.~Charlin, and
  T.~Tuytelaars.
\newblock Online continual learning with maximally interfered retrieval.
\newblock \emph{arXiv preprint arXiv:1908.04742}, 2019{\natexlab{a}}.

\bibitem[Aljundi et~al.(2019{\natexlab{b}})Aljundi, Lin, Goujaud, and
  Bengio]{aljundi2019gradient}
R.~Aljundi, M.~Lin, B.~Goujaud, and Y.~Bengio.
\newblock Gradient based sample selection for online continual learning.
\newblock \emph{NeurIPS}, 32, 2019{\natexlab{b}}.

\bibitem[Bang et~al.(2022)Bang, Koh, Park, Song, Ha, and Choi]{bang2022online}
J.~Bang, H.~Koh, S.~Park, H.~Song, J.-W. Ha, and J.~Choi.
\newblock Online continual learning on a contaminated data stream with blurry
  task boundaries.
\newblock In \emph{CVPR}, pages 9275--9284, 2022.

\bibitem[Bendale and Boult(2015)]{bendale2015towards}
A.~Bendale and T.~Boult.
\newblock Towards open world recognition.
\newblock In \emph{CVPR}, pages 1893--1902, 2015.

\bibitem[Caron et~al.(2021)Caron, Touvron, Misra, J\'egou, Mairal, Bojanowski,
  and Joulin]{caron2021emerging}
M.~Caron, H.~Touvron, I.~Misra, H.~J\'egou, J.~Mairal, P.~Bojanowski, and
  A.~Joulin.
\newblock Emerging properties in self-supervised vision transformers.
\newblock In \emph{ICCV}, 2021.

\bibitem[Chaudhry et~al.(2019{\natexlab{a}})Chaudhry, Ranzato, Rohrbach, and
  Elhoseiny]{chaudhry2019efficient}
A.~Chaudhry, M.~Ranzato, M.~Rohrbach, and M.~Elhoseiny.
\newblock Efficient lifelong learning with a-gem.
\newblock In \emph{ICLR}, 2019{\natexlab{a}}.

\bibitem[Chaudhry et~al.(2019{\natexlab{b}})Chaudhry, Rohrbach, Elhoseiny,
  Ajanthan, Dokania, Torr, and Ranzato]{chaudhry2019continual}
A.~Chaudhry, M.~Rohrbach, M.~Elhoseiny, T.~Ajanthan, P.~K. Dokania, P.~H. Torr,
  and M.~Ranzato.
\newblock Continual learning with tiny episodic memories.
\newblock \emph{ICML}, 2019{\natexlab{b}}.

\bibitem[De~Lange et~al.(2021)De~Lange, Aljundi, Masana, Parisot, Jia,
  Leonardis, Slabaugh, and Tuytelaars]{de2021continual}
M.~De~Lange, R.~Aljundi, M.~Masana, S.~Parisot, X.~Jia, A.~Leonardis,
  G.~Slabaugh, and T.~Tuytelaars.
\newblock A continual learning survey: Defying forgetting in classification
  tasks.
\newblock \emph{IEEE TPAMI}, 44\penalty0 (7):\penalty0 3366--3385, 2021.

\bibitem[Fei and Liu(2016)]{fei2016breaking}
G.~Fei and B.~Liu.
\newblock Breaking the closed world assumption in text classification.
\newblock In \emph{Proceedings of the 2016 Conference of the North American
  Chapter of the Association for Computational Linguistics: Human Language
  Technologies}, pages 506--514, 2016.

\bibitem[Fei et~al.(2016)Fei, Wang, and Liu]{fei2016learning}
G.~Fei, S.~Wang, and B.~Liu.
\newblock Learning cumulatively to become more knowledgeable.
\newblock In \emph{KDD-2026}, pages 1565--1574, 2016.

\bibitem[Fisher(1936)]{fisher1936use}
R.~A. Fisher.
\newblock The use of multiple measurements in taxonomic problems.
\newblock \emph{Annals of eugenics}, 7\penalty0 (2):\penalty0 179--188, 1936.

\bibitem[Ghassemi and Fazl-Ersi(2022)]{ghassemi2022comprehensive}
N.~Ghassemi and E.~Fazl-Ersi.
\newblock A comprehensive review of trends, applications and challenges in
  out-of-distribution detection.
\newblock \emph{arXiv preprint arXiv:2209.12935}, 2022.

\bibitem[Gummadi et~al.(2022)Gummadi, Kent, Mendez, and
  Eaton]{gummadi2022shels}
M.~Gummadi, D.~Kent, J.~A. Mendez, and E.~Eaton.
\newblock Shels: Exclusive feature sets for novelty detection and continual
  learning without class boundaries.
\newblock In \emph{CoLLaS}, pages 1065--1085. PMLR, 2022.

\bibitem[Han et~al.(2021)Han, Rebuffi, Ehrhardt, Vedaldi, and
  Zisserman]{han2021autonovel}
K.~Han, S.-A. Rebuffi, S.~Ehrhardt, A.~Vedaldi, and A.~Zisserman.
\newblock Autonovel: Automatically discovering and learning novel visual
  categories.
\newblock \emph{IEEE TPAMI}, 44\penalty0 (10):\penalty0 6767--6781, 2021.

\bibitem[Hayes and Kanan(2020)]{hayes2020lifelong}
T.~L. Hayes and C.~Kanan.
\newblock Lifelong machine learning with deep streaming linear discriminant
  analysis.
\newblock In \emph{CVPR workshops}, 2020.

\bibitem[He and Zhu(2022)]{he2022out}
J.~He and F.~Zhu.
\newblock Out-of-distribution detection in unsupervised continual learning.
\newblock In \emph{CVPR}, pages 3850--3855, 2022.

\bibitem[Hendrycks and Gimpel(2017)]{hendrycks2017baseline}
D.~Hendrycks and K.~Gimpel.
\newblock A baseline for detecting misclassified and out-of-distribution
  examples in neural networks.
\newblock In \emph{ICLR}, 2017.

\bibitem[Hendrycks et~al.(2018)Hendrycks, Mazeika, and
  Dietterich]{hendrycks2018deep}
D.~Hendrycks, M.~Mazeika, and T.~Dietterich.
\newblock Deep anomaly detection with outlier exposure.
\newblock \emph{arXiv preprint arXiv:1812.04606}, 2018.

\bibitem[Huang and Li(2021)]{huang2021mos}
R.~Huang and Y.~Li.
\newblock Mos: Towards scaling out-of-distribution detection for large semantic
  space.
\newblock In \emph{CVPR}, pages 8710--8719, 2021.

\bibitem[Kaymak et~al.(2025)Kaymak, Kim, Kaichi, Konishi, and Liu]{PLDA_github}
D.~Kaymak, G.~Kim, T.~Kaichi, T.~Konishi, and B.~Liu.
\newblock {PLDA}.
\newblock \url{https://github.com/drdkymk/PLDA}, 2025.
\newblock URL \url{https://github.com/drdkymk/PLDA}.
\newblock GitHub repository.

\bibitem[Ke and Liu(2022)]{ke2022continualsurvey}
Z.~Ke and B.~Liu.
\newblock Continual learning of natural language processing tasks: A survey.
\newblock \emph{arXiv preprint arXiv:2211.12701}, 2022.

\bibitem[Kemker and Kanan(2018)]{Kemker2018fearnet}
R.~Kemker and C.~Kanan.
\newblock {FearNet: Brain-Inspired Model for Incremental Learning}.
\newblock In \emph{ICLR}, 2018.

\bibitem[Kim et~al.(2022{\natexlab{a}})Kim, Liu, and Ke]{kim2022multi}
G.~Kim, B.~Liu, and Z.~Ke.
\newblock A multi-head model for continual learning via out-of-distribution
  replay.
\newblock In \emph{CoLLaS}, pages 548--563, 2022{\natexlab{a}}.

\bibitem[Kim et~al.(2022{\natexlab{b}})Kim, Xiao, Konishi, Ke, and
  Liu]{kim2022theoretical}
G.~Kim, C.~Xiao, T.~Konishi, Z.~Ke, and B.~Liu.
\newblock A theoretical study on solving continual learning.
\newblock \emph{NeurIPS}, 35:\penalty0 5065--5079, 2022{\natexlab{b}}.

\bibitem[Kim et~al.(2025)Kim, Xiao, Konishi, Ke, and Liu]{kim2025open}
G.~Kim, C.~Xiao, T.~Konishi, Z.~Ke, and B.~Liu.
\newblock Open-world continual learning: Unifying novelty detection and
  continual learning.
\newblock \emph{Artificial Intelligence}, 338:\penalty0 104237, 2025.

\bibitem[Kim et~al.(2021)Kim, Koo, and Hwang]{kim2021unified}
J.~Kim, J.~Koo, and S.~Hwang.
\newblock A unified benchmark for the unknown detection capability of deep
  neural networks.
\newblock \emph{arXiv preprint arXiv:2112.00337}, 2021.

\bibitem[Kirkpatrick et~al.(2017)Kirkpatrick, Pascanu, Rabinowitz, Veness,
  Desjardins, Rusu, Milan, Quan, Ramalho, Grabska-Barwinska,
  et~al.]{Kirkpatrick2017overcoming}
J.~Kirkpatrick, R.~Pascanu, N.~Rabinowitz, J.~Veness, G.~Desjardins, A.~A.
  Rusu, K.~Milan, J.~Quan, T.~Ramalho, A.~Grabska-Barwinska, et~al.
\newblock Overcoming catastrophic forgetting in neural networks.
\newblock \emph{Proceedings of the national academy of sciences}, 114\penalty0
  (13):\penalty0 3521--3526, 2017.

\bibitem[Koh et~al.(2022)Koh, Kim, Ha, and Choi]{koh2022online}
H.~Koh, D.~Kim, J.-W. Ha, and J.~Choi.
\newblock Online continual learning on class incremental blurry task
  configuration with anytime inference.
\newblock In \emph{ICLR}, 2022.

\bibitem[Krizhevsky et~al.(2009)Krizhevsky, Hinton,
  et~al.]{krizhevsky2009learning}
A.~Krizhevsky, G.~Hinton, et~al.
\newblock Learning multiple layers of features from tiny images.
\newblock 2009.

\bibitem[Lakshminarayanan et~al.(2017)Lakshminarayanan, Pritzel, and
  Blundell]{lakshminarayanan2017simple}
B.~Lakshminarayanan, A.~Pritzel, and C.~Blundell.
\newblock Simple and scalable predictive uncertainty estimation using deep
  ensembles.
\newblock \emph{NeurIPS}, 2017.

\bibitem[Le and Yang(2015)]{le2015tiny}
Y.~Le and X.~Yang.
\newblock Tiny imagenet visual recognition challenge.
\newblock \emph{CS 231N}, 7\penalty0 (7):\penalty0 3, 2015.

\bibitem[Lee et~al.(2018)Lee, Lee, Lee, and Shin]{lee2018simple}
K.~Lee, K.~Lee, H.~Lee, and J.~Shin.
\newblock A simple unified framework for detecting out-of-distribution samples
  and adversarial attacks.
\newblock \emph{NeurIPS}, 2018.

\bibitem[Li and Hoiem(2016)]{Li2016LwF}
Z.~Li and D.~Hoiem.
\newblock {Learning Without Forgetting}.
\newblock In \emph{ECCV}, 2016.

\bibitem[Lin et~al.(2022)Lin, Yang, Fan, and Zhang]{lin2022beyond}
S.~Lin, L.~Yang, D.~Fan, and J.~Zhang.
\newblock Beyond not-forgetting: Continual learning with backward knowledge
  transfer.
\newblock \emph{NeurIPS}, 2022.

\bibitem[Liu et~al.(2023{\natexlab{a}})Liu, Mazumder, Robertson, and
  Grigsby]{liu2023ai}
B.~Liu, S.~Mazumder, E.~Robertson, and S.~Grigsby.
\newblock Ai autonomy: Self-initiated open-world continual learning and
  adaptation.
\newblock \emph{AI Magazine}, 2023{\natexlab{a}}.

\bibitem[Liu et~al.(2023{\natexlab{b}})Liu, Lochman, and Zach]{liu2023gen}
X.~Liu, Y.~Lochman, and C.~Zach.
\newblock Gen: Pushing the limits of softmax-based out-of-distribution
  detection.
\newblock In \emph{CVPR}, 2023{\natexlab{b}}.

\bibitem[Mai et~al.(2021)Mai, Li, Kim, and Sanner]{mai2021supervised}
Z.~Mai, R.~Li, H.~Kim, and S.~Sanner.
\newblock Supervised contrastive replay: Revisiting the nearest class mean
  classifier in online class-incremental continual learning.
\newblock In \emph{CVPR Workshops}, pages 3589--3599, 2021.

\bibitem[Mai et~al.(2022)Mai, Li, Jeong, Quispe, Kim, and
  Sanner]{mai2022online}
Z.~Mai, R.~Li, J.~Jeong, D.~Quispe, H.~Kim, and S.~Sanner.
\newblock Online continual learning in image classification: An empirical
  survey.
\newblock \emph{Neurocomputing}, 469:\penalty0 28--51, 2022.

\bibitem[Malinin et~al.(2019)Malinin, Mlodozeniec, and
  Gales]{malinin2019ensemble}
A.~Malinin, B.~Mlodozeniec, and M.~Gales.
\newblock Ensemble distribution distillation.
\newblock In \emph{ICLR}, 2019.

\bibitem[Mallya and Lazebnik(2017)]{Mallya2017packnet}
A.~Mallya and S.~Lazebnik.
\newblock {PackNet: Adding Multiple Tasks to a Single Network by Iterative
  Pruning}.
\newblock \emph{arXiv preprint arXiv:1711.05769}, 2017.

\bibitem[McLachlan(1999)]{mclachlan1999mahalanobis}
G.~J. McLachlan.
\newblock Mahalanobis distance.
\newblock \emph{Resonance}, 4\penalty0 (6):\penalty0 20--26, 1999.

\bibitem[Mi et~al.(2020)Mi, Kong, Lin, Yu, and Faltings]{mi2020generalized}
F.~Mi, L.~Kong, T.~Lin, K.~Yu, and B.~Faltings.
\newblock Generalized class incremental learning.
\newblock In \emph{CVPR workshops}, pages 240--241, 2020.

\bibitem[Momeni et~al.(2025)Momeni, Mazumder, and Liu]{momeni2025continual}
S.~Momeni, S.~Mazumder, and B.~Liu.
\newblock Continual learning using a kernel-based method over foundation
  models.
\newblock In \emph{AAAI-2025}, volume~39, pages 19528--19536, 2025.

\bibitem[Oquab et~al.(2023)Oquab, Darcet, Moutakanni, Vo, Szafraniec, Khalidov,
  Fernandez, Haziza, Massa, El-Nouby, et~al.]{oquab2023dinov2}
M.~Oquab, T.~Darcet, T.~Moutakanni, H.~Vo, M.~Szafraniec, V.~Khalidov,
  P.~Fernandez, D.~Haziza, F.~Massa, A.~El-Nouby, et~al.
\newblock Dinov2: Learning robust visual features without supervision.
\newblock \emph{arXiv preprint arXiv:2304.07193}, 2023.

\bibitem[Pang et~al.(2005)Pang, Ozawa, and Kasabov]{pang2005incremental}
S.~Pang, S.~Ozawa, and N.~Kasabov.
\newblock Incremental linear discriminant analysis for classification of data
  streams.
\newblock \emph{IEEE transactions on systems, man, and cybernetics},
  35:\penalty0 905--14, 11 2005.

\bibitem[Papadopoulos et~al.(2021)Papadopoulos, Rajati, Shaikh, and
  Wang]{papadopoulos2021outlier}
A.-A. Papadopoulos, M.~R. Rajati, N.~Shaikh, and J.~Wang.
\newblock Outlier exposure with confidence control for out-of-distribution
  detection.
\newblock \emph{Neurocomputing}, 441:\penalty0 138--150, 2021.

\bibitem[Prabhu et~al.(2020)Prabhu, Torr, and Dokania]{prabhu2020gdumb}
A.~Prabhu, P.~H. Torr, and P.~K. Dokania.
\newblock Gdumb: A simple approach that questions our progress in continual
  learning.
\newblock In \emph{EECV}, 2020.

\bibitem[Rebuffi et~al.(2017)Rebuffi, Kolesnikov, Sperl, and
  Lampert]{rebuffi2017icarl}
S.-A. Rebuffi, A.~Kolesnikov, G.~Sperl, and C.~H. Lampert.
\newblock icarl: Incremental classifier and representation learning.
\newblock In \emph{CVPR}, 2017.

\bibitem[Ren et~al.(2021)Ren, Fort, Liu, Roy, Padhy, and
  Lakshminarayanan]{ren2021simple}
J.~Ren, S.~Fort, J.~Liu, A.~G. Roy, S.~Padhy, and B.~Lakshminarayanan.
\newblock A simple fix to mahalanobis distance for improving near-ood
  detection.
\newblock \emph{arXiv preprint arXiv:2106.09022}, 2021.

\bibitem[Rios et~al.(2022)Rios, Ahuja, Ndiour, Genc, Itti, and
  Tickoo]{rios2022incdfm}
A.~Rios, N.~Ahuja, I.~Ndiour, U.~Genc, L.~Itti, and O.~Tickoo.
\newblock incdfm: Incremental deep feature modeling for continual novelty
  detection.
\newblock In \emph{ECCV 2022}, 2022.

\bibitem[Roy et~al.(2022)Roy, Liu, Zhong, Sebe, and Ricci]{roy2022class}
S.~Roy, M.~Liu, Z.~Zhong, N.~Sebe, and E.~Ricci.
\newblock Class-incremental novel class discovery.
\newblock In \emph{ECCV}, pages 317--333, 2022.

\bibitem[Serra et~al.(2018)Serra, Suris, Miron, and
  Karatzoglou]{Serra2018overcoming}
J.~Serra, D.~Suris, M.~Miron, and A.~Karatzoglou.
\newblock Overcoming catastrophic forgetting with hard attention to the task.
\newblock In \emph{ICML}, 2018.

\bibitem[Sun and Li(2022)]{sun2022dice}
Y.~Sun and Y.~Li.
\newblock Dice: Leveraging sparsification for out-of-distribution detection.
\newblock In \emph{ECCV}, 2022.

\bibitem[Touvron et~al.(2021)Touvron, Cord, Douze, Massa, Sablayrolles, and
  J{\'e}gou]{touvron2021training}
H.~Touvron, M.~Cord, M.~Douze, F.~Massa, A.~Sablayrolles, and H.~J{\'e}gou.
\newblock Training data-efficient image transformers \& distillation through
  attention.
\newblock In \emph{ICML}, pages 10347--10357. PMLR, 2021.

\bibitem[Van~de Ven and Tolias(2019)]{van2019three}
G.~M. Van~de Ven and A.~S. Tolias.
\newblock Three scenarios for continual learning.
\newblock \emph{arXiv preprint arXiv:1904.07734}, 2019.

\bibitem[Wang et~al.(2023)Wang, Zhang, Su, and Zhu]{wang2023comprehensive}
L.~Wang, X.~Zhang, H.~Su, and J.~Zhu.
\newblock A comprehensive survey of continual learning: Theory, method and
  application, 2023.

\bibitem[Wortsman et~al.(2020)Wortsman, Ramanujan, Liu, Kembhavi, Rastegari,
  Yosinski, and Farhadi]{wortsman2020supermasks}
M.~Wortsman, V.~Ramanujan, R.~Liu, A.~Kembhavi, M.~Rastegari, J.~Yosinski, and
  A.~Farhadi.
\newblock Supermasks in superposition.
\newblock \emph{NeurIPS}, 2020.

\bibitem[Wu et~al.(2018)Wu, Herranz, Liu, van~de Weijer, Raducanu,
  et~al.]{wu2018memory}
C.~Wu, L.~Herranz, X.~Liu, J.~van~de Weijer, B.~Raducanu, et~al.
\newblock Memory replay gans: Learning to generate new categories without
  forgetting.
\newblock In \emph{NIPS}, pages 5962--5972, 2018.

\bibitem[Yang et~al.(2021)Yang, Zhou, Li, and Liu]{yang2021generalized}
J.~Yang, K.~Zhou, Y.~Li, and Z.~Liu.
\newblock Generalized out-of-distribution detection: A survey.
\newblock \emph{arXiv preprint arXiv:2110.11334}, 2021.

\bibitem[Zeng et~al.(2019)Zeng, Chen, Cui, and Yu]{zeng2019continuous}
G.~Zeng, Y.~Chen, B.~Cui, and S.~Yu.
\newblock Continuous learning of context-dependent processing in neural
  networks.
\newblock \emph{Nature Machine Intelligence}, 2019.

\bibitem[Zhuang et~al.(2024)Zhuang, Chen, Fang, He, Tong, Wei, Zeng, and
  Chen]{zhuang2024gacl}
H.~Zhuang, Y.~Chen, D.~Fang, R.~He, K.~Tong, H.~Wei, Z.~Zeng, and C.~Chen.
\newblock {GACL}: Exemplar-free generalized analytic continual learning.
\newblock In \emph{NeurIPS}, 2024.

\end{thebibliography}

\end{document}